\def\year{2020}\relax

\documentclass[letterpaper]{article} 
\usepackage{aaai20}  
\usepackage{times}  
\usepackage{helvet} 
\usepackage{courier}  
\usepackage[hyphens]{url}  
\usepackage{graphicx} 
\usepackage{graphicx}  
\nocopyright
\title{Few-Shot Learning for Road Object Detection}
\author{
\Large \textbf{
Anay Majee\thanks{Equal contribution},
Kshitij Agrawal\footnotemark[1],
Anbumani Subramanian
} \\ 
Intel Corporation\\ 
\{anay.majee, kshitij.agrawal, anbumani.subramanian\}@intel.com
}

\begin{document}
\maketitle

\begin{abstract}
Few-shot learning is a problem of high interest in the evolution of deep learning. In this work, we consider the problem of few-shot object detection (FSOD) in a real-world, class-imbalanced scenario. For our experiments, we utilize the India Driving Dataset (IDD), as it includes a class of less-occurring road objects in the image dataset and hence provides a setup suitable for few-shot learning. We evaluate both metric-learning and meta-learning based FSOD methods, in two experimental settings: (i) representative (same-domain) splits from IDD, that evaluates the ability of a model to learn in the context of road images, and (ii) object classes with less-occurring object samples, similar to the open-set setting in real-world. From our experiments, we demonstrate that the metric-learning method outperforms meta-learning on the novel classes by (i) 11.2 $mAP$ points on the same domain, and (ii) 1.0 $mAP$ point on the open-set. We also show that our extension of object classes in a real-world open dataset offers a rich ground for few-shot learning studies.
\end{abstract}

\section{Introduction}
Researchers have long aimed to replicate the innate ability of humans to identify representative features of an object from few examples and relate this knowledge to learning of new objects, even with limited examples. 

Convolutional neural networks (CNNs) have made great progress in the fields of object classification \cite{resnet} and detection \cite{yolo1,ssd,faster-rcnn}. These methods rely on datasets with labelled samples, in the order of millions of images to achieve high accuracy. This in turn entails a high human cost in capturing and labelling a large distribution of images. The cost of training and retraining such models also increases as the dataset size increases. Few-shot learning is a paradigm of machine learning that relies on only a few examples of labelled data to train, unlike the traditional approach for classification or object detection which requires thousands of samples per object type.

\begin{figure}[t]
\centering
\begin{tabular}{ccc}
\includegraphics[height=0.21\linewidth,width=0.30\linewidth]{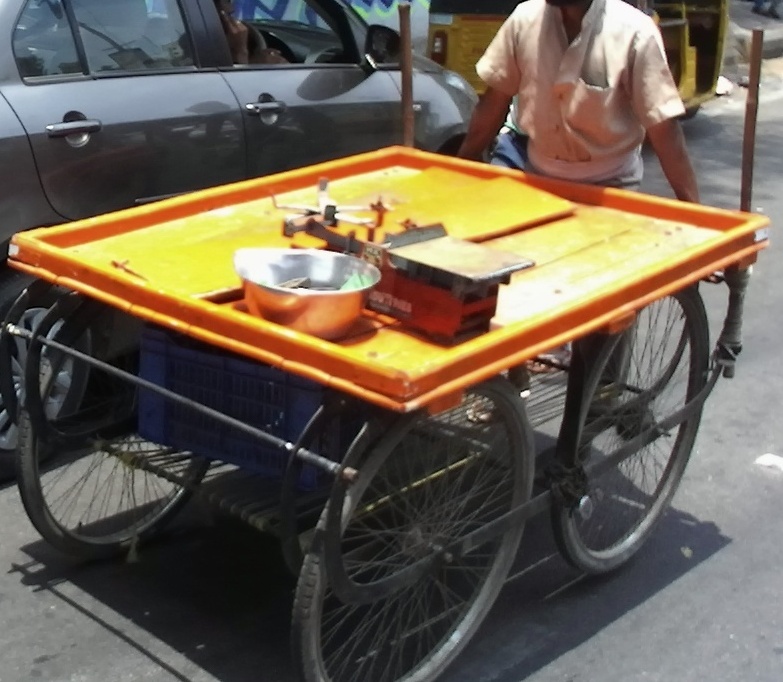}
&
\includegraphics[height=0.21\linewidth,width=0.30\linewidth]{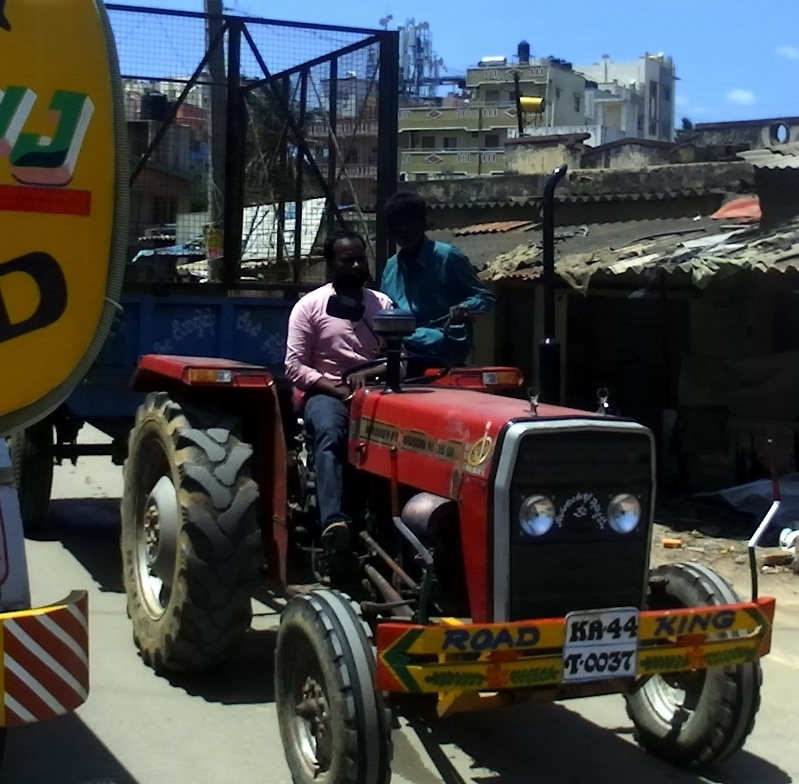}
&
\includegraphics[height=0.21\linewidth,width=0.30\linewidth]{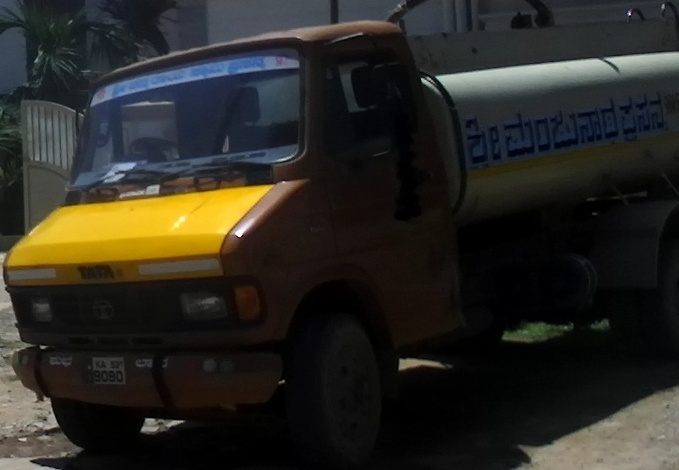}\\
\end{tabular} 
\begin{tabular}{cc}
\includegraphics[height=0.21\linewidth,width=0.30\linewidth]{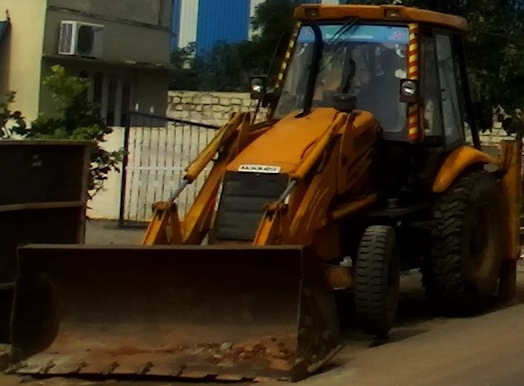}
&
\includegraphics[height=0.21\linewidth,width=0.30\linewidth]{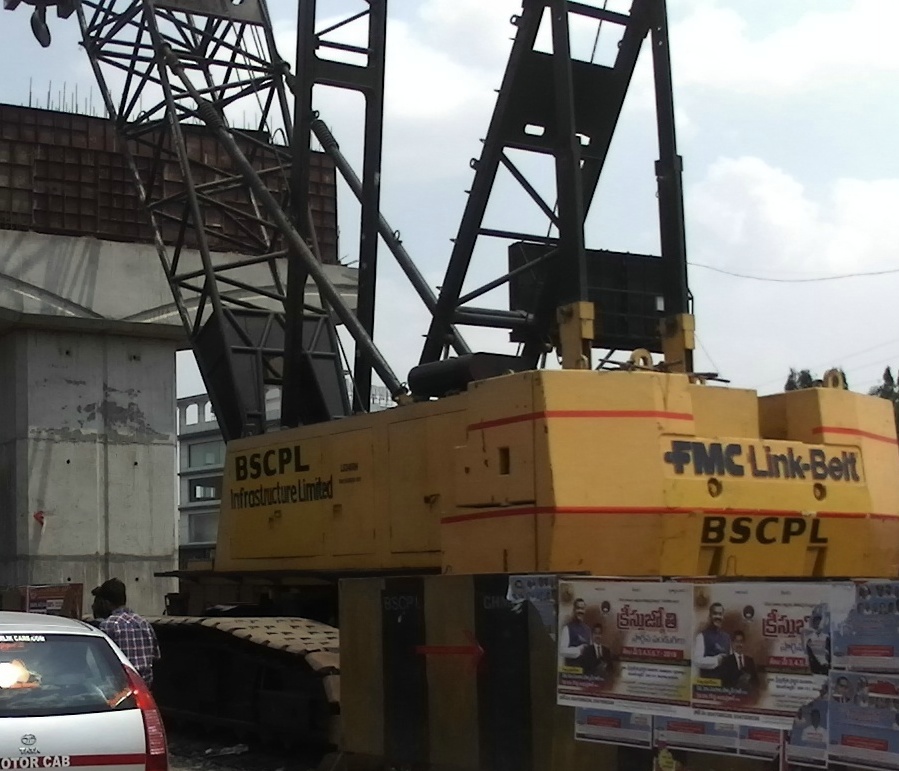}
\end{tabular}
\caption{Sample images of less-occurring road objects from the $vehicle$-$fallback$ class of IDD. Some new object categories identified in this class include $street$-$cart$, $tractor$, $water$-$tanker$, $excavator$, and $crane$.}
\label{fig_vfb}
\end{figure}

Few-shot learning has been successfully used in the classification task \cite{matching-net,protonet,maml} with the introduction of approaches like prototype generation and meta-learning by adding an auxiliary network to help in learning to learn.  The best performing approach is based on a cosine-based distance classifier \cite{matching-net}. Few-shot object detection networks borrow from classification approaches and implement them in one stage \cite{reweight} or two-stage \cite{metarcnn,addfeat,fsdet} detection.

Many existing few-shot object detection methods are trained and optimized on canonical datasets like PASCAL-VOC \cite{voc} or MS-COCO \cite{coco}. However, none of these methods has explored datasets with  images from real-world scenarios, like driving datasets. The context of driving is unique as it provides a rich visual scene with varying densities of objects in images. Moreover, due to the dynamic and open nature of the world, we observe a long tail distribution of categories that inherently contains class-imbalance. We consider this an important aspect to be considered, especially for real-world deployments. Towards this end, we select the India Driving Dataset (IDD) \cite{idd}. Figure \ref{fig_vfb} shows examples of some new object categories identified in the existing $vehicle-fallback$ category in IDD.

In this work, we focus on the problem of few-shot object detection, and extend the capability of a trained object detector with base-classes to novel classes for road object detection. Our contributions include:
\begin{enumerate}
    \item demonstration of few-shot learning approach in real-world images from a driving context 
    \item extension of an existing object class with less-occurring object samples in an open dataset to  show the viability and efficacy of learning from a few samples, and
    \item analysis of metric vs. meta-learning methods in a real-world object detection problem.
\end{enumerate}

To the best of our knowledge, none of the existing works demonstrate few-shot learning on real-world images in a driving scenario which we demonstrate through our work.

The paper is organized as follows: sec. \ref{related} covers related works on Few-Shot learning and Object Detection, sec. \ref{approach} highlights the various approaches applied to solve the challenge of Few-Shot Object Detection and sec. \ref{experiment} details the experiments conducted on few-shot data splits of IDD dataset and the findings from these experiments.

\section{Related Work}
\label{related}
\subsection{Few-Shot Learning}
\label{related:fscl}
Few-shot learning is a technique that learns data distributions based on only a few examples of a particular category. It is an active area of research in computer vision and machine learning \cite{closerfewshot}.
The objective of few-shot learning is to learn a generalizable feature extractor and adapt to the novel (or new) classes containing only a few - say 10 to 50 - samples \cite{protonet,matching-net}, in a $N$-way, $K$-shot fashion.
Most of the techniques in this area employ episodic training strategies \cite{local_desc,matching-net,protonet} to guide a learning algorithm to adapt given limited data samples.
An episode is formed by sampling a small subset of images from a large scale image dataset like ImageNet \cite{imagenet} or MS-COCO \cite{coco}. Each subset consists of $N$ classes with $K$ samples, referred to as \emph{support set}, and $Q$ examples per class in the evaluation or fine-tuning stage which is the \emph{query set}.   
A  promising set of techniques in this direction are built on meta-learning \cite{maml,local_desc,relation-net,closerfewshot} and differ by how they encode the knowledge from the episodic training and utilize this difference to boost performance on the novel classes.

A prominent limitation of these techniques is $catastrophic$ $forgetting$ wherein the network forgets the already learnt categories (also called base-category) while learning the newly added ones (also called novel-category). Few techniques employ the use of non-linear feature similarity estimation techniques \cite{nonforget} to address this challenge. 

The above works tackle the problem of image classification, whereas our work focuses on object detection.

\subsection{Few-Shot Object Detection}
\label{related:fsod}
Few-shot object detection draws inspiration from the classification setting and extends the capabilities of existing object detection networks \cite{yolo2,faster-rcnn} to adapt to fewer data samples. General object detection can be divided into $proposal$-$free$ \cite{yolo2,retinanet,polyyolo} and $proposal$-$based$ \cite{faster-rcnn,fpn} architectures which rely on large scale detection datasets to learn a distribution with high accuracy. Few-shot architectures on the other hand have to learn a similar distribution with very limited data.

Initial attempts in few-shot object detection have used transfer learning \cite{lstd} and distance-metric learners \cite{repmet} to adapt to the novel classes, given a generalizable feature extractor trained on abundant data samples from the base classes. More recent literature makes use of non-linear similarity operators like cosine similarity \cite{fsdet} to learn distinguishable features between base and novel classes. Some use class-specific attention vectors to re-weight full image features as in YOLO \cite{reweight}, or Region of Interest (RoI) features as in Faster-RCNN \cite{metarcnn}, while \cite{fgn,fsod} use attentive vectors along with relational operators \cite{relation-net} to learn low-level features that distinguish the base categories from the novel ones. In \cite{addfeat}, the authors demonstrate the use of additional features to further guide the object detection network. 
Along this line, \cite{metadet} separates the learning between category-specific and category-agnostic components and points out the issue of confusion between categories in the classification of predicted ROIs as a limitation of existing few-shot object detection networks. 

Our work evaluates various techniques in the domain of few-shot object detection to predict the occurrence of rare and unidentified road objects (only a few examples per category are available) in unstructured driving conditions.

\section{Approach}
\label{approach}
\begin{figure*}[t]
      \centering
      \includegraphics[width=0.85\textwidth]{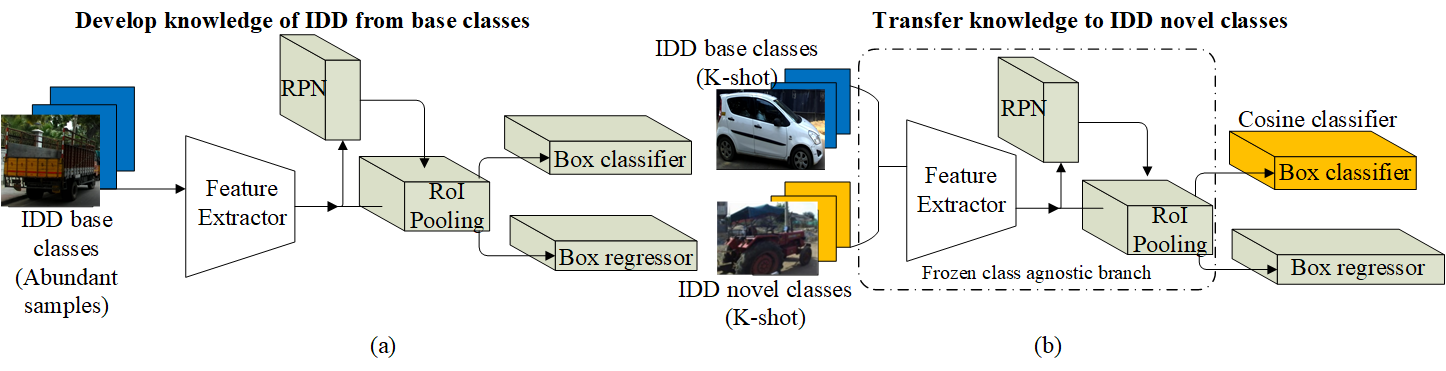}
      \caption{Two-stage fine-tuning architecture: (a) base training stage where the network learns from abundant samples from the base dataset, and (b) fine-tuning stage where the network adapts to few-shot samples.}
      \label{fig_tfa}
\end{figure*}

\subsection{Problem Definition}
\label{approach:prob_def}
Following the formulation of the few-shot learning problem in \cite{reweight,metarcnn}, we define a few shot learner, represented by $h(I,\theta)$ that receives input data ($I$) from base classes $C_{base}$ and novel classes $C_{novel}$. Here, $\theta$ denotes the learnable parameters. The training data can be divided into two distinct parts:
\begin{itemize}
      \item $D_{base} = {(x_{i}^{base}, y_{i}^{base})}_{i=1}^{b}$ containing abundant training examples from $b$ base classes.
      \item $D_{novel} = {(x_{i}^{novel}, y_{i}^{novel})}_{i=1}^{n}$ containing only $K$-shot training examples from $n$ novel classes.
\end{itemize}
The objective of the few-shot learner is to learn a generalizable feature representation from $D_{base}$ and transfer the knowledge to the novel classes in $D_{novel}$, such that the performance of $h(I,\theta)$ increases on the novel classes with minimal degradation on the base classes. 

From our literature review, we adopt a two-stage training mechanism and conduct experiments on multiple few-shot architectures which can be categorised broadly into two categories: 
\begin{itemize}
      \item \textbf{Feature similarity based} - The first category following \cite{fsdet} uses abundant data samples from $D_{base}$ in the first stage and adopts a fine-tuning mechanism using $K$-shot data samples from $D_{base} \cup D_{novel}$. During fine-tuning, it learns a distance metric to learn distinguishing features between base and novel classes.
      \item \textbf{Auxiliary network based} - The second follows \cite{reweight,addfeat,metarcnn} and splits the training dataset into episodes where each episode is designed as a task $T_{j}$. A task or an episode simulates the real-world scenario for a few-shot learner and allows $h(I,\theta)$ to adapt to few-shot data. We define $T_{j} = S_{j} \cup Q_{j}$ where $S$ and $Q$ represents support and query sets respectively. In every training iteration, $h(I,\theta)$ trains on the support set and is evaluated/fine-tuned on the query set (loss is computed on this set). The support set samples images from $D_{base}$ in the first stage and from $D_{base} \cup D_{novel}$ in the second stage.
\end{itemize}

The learnt model $h(I,\theta)$ is evaluated on $D_{test}$ containing unseen data samples from both $C_{base}$ and $C_{novel}$. The learning objective of $h(I,\theta)$ becomes difficult in case of object detection task, specifically in driving scenarios where multiple objects (ROIs) are present on the same input image $x_{i}$ and in scenarios with occlusion. 

\begin{figure}[t]
\centering  
\includegraphics[width=0.9\columnwidth,height=4cm]{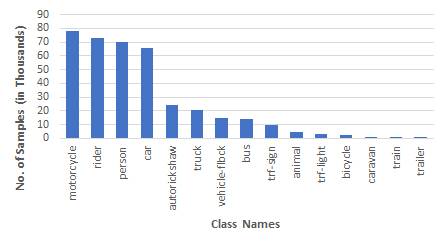}
\caption{Class-wise distribution from IDD-Detection in order of highest to lowest. We can observe the long tail distribution of data from this figure. Vehicle-fallback class is selected for labelling new samples.}
\label{fig_dist}
\end{figure}

\begin{figure*}[t]
      \centering
      \includegraphics[width=0.75\textwidth,keepaspectratio]{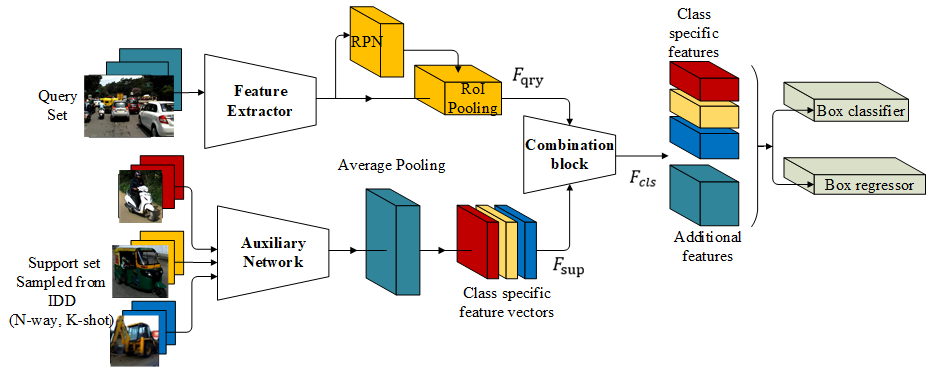}
      \caption{Meta-learning based Few-shot object detection architecture as used in our work. Adapted from the training strategies mentioned in \cite{metarcnn,addfeat}.}
      \label{fig_aux}
\end{figure*}

\subsection{Datasets}
\label{approach:data}
Object detection tasks are commonly benchmarked on canonical datasets like PASCAL-VOC or MS-COCO. While these datasets provide a representative indication of performance they are artificially compiled with a balanced class distribution. Most real-world systems do not contain such balanced visual representations of every object.  We observe that real-world often has a long-tail representation with very few examples of such categories. Thus a training of object detection methods does not translate well to real-world scenarios like in driving. Towards this end, we considered  using driving datasets which provide examples of objects in a real-world context. Some popular datasets for driving tasks are KITTI \cite{kitti}, Cityscapes \cite{city} and India Driving Dataset (IDD) \cite{idd}. 

\begin{table} [t]
    \caption{Distribution of new-categories after expanding vehicle-fallback class from IDD-Detection considering only objects of size greater than 100x100 pixels.}
    \centering
    \begin{tabular}{ c | c} 
          \hline
          Category          & No. of instances \\
          \hline
          Street carts      & 221 \\
          Tractor           & 89 \\
          Water tanker      & 42 \\
          Excavator         & 43 \\
          Crane             & 20 \\
          \hline
          Total validation images &242\footnotemark\\
          \hline
    \end{tabular}
    \label{tableI}
\end{table}
\footnotetext{Re-classified from train and validation set of IDD.}

Of these, KITTI provides only 3 types of 2D object annotations, Cityscapes does not contain any bounding box annotation and provides 18 classes for the task of semantic segmentation. Cityscapes dataset has been extended to 3D bounding boxes for vehicles, however, it is not relevant to our task of 2D object detection. The IDD-Detection dataset\footnote{https://idd.insaan.iiit.ac.in/dataset/details/} consists of 30 object categories with 26 trainable categories for fine-grained semantic segmentation and 15 categories for 2D object detection. For training few-shot methods, we require to partition our dataset into base and novel classes. Additionally, we want to evaluate the few-shot techniques in an open-set setting for which we require the ability to add low-occurrence classes for detection. We selected the IDD-Detection Dataset given two aspects: (i) high class-imbalance like in real-world and high variability in the type of road objects, and (ii) the possibility to extend classes for detection. From Fig. \ref{fig_dist} we can see the severe class imbalance in the distribution of classes. 

To extend the learning for fine-grained detection, we rely on the existing  \emph{vehicle-fallback} class in IDD. As described by the authors \cite{idd}, the vehicle-fallback class has been created as a place-holder for vehicle categories that do not fall into any of the other predefined 19 object categories. We have manually categorized samples from the \emph{vehicle-fallback} class for further expansion into 5 new categories\footnote{\url{https://github.com/intel/driving-data-collection-reference-kit/tree/master/few-shot-annotations-AAAI-21}} -  street-cart, tractor and water-tanker, excavator, crane. In Table \ref{tableI} we observe that the number of total samples of these novel classes is very few and hence closely resembles the open-world challenge. Due to the limited number of samples, these categories are ideal for few-shot learning. Further, for evaluating few-shot learning algorithms on IDD we utilize the complete validation set of IDD consisting of 10225 images from all 15 categories.

\subsection{Few-shot Tasks}
\label{approach:data_split}
For training the few-shot detectors we define two task settings based on the categorization of data, namely same domain (IDD-10) and open-set (IDD-OS).
\begin{enumerate}
    \item \textbf{Same domain (IDD-10)}: From the IDD-Detection dataset of 15 classes, we consider only 10 and discard the following - \emph{animal, caravan, train, trailer} and \emph{vehicle-fallback}. These classes are either incorrectly labelled (\textit{caravan, train, trailer}) or have very high variability in representation (different animals all labeled as \textit{animals}). From the remaining 10 classes, we create two splits with 7 base-classes trained in the base-training stage and 3 novel classes trained in the fine-tuning stage. This is done based on sampling the distribution to include high variability representations in test phase. The two novel splits for our experiments are - (i) $bicycle$, $bus$ and $truck$, (ii) $autorickshaw$, $motorcycle$ and $truck$. 
    \item \textbf{Open-Set (IDD-OS)}: This setting effectively represents a real-world deployment setting where a system will be trained to recognize new classes using few-shot learning. To simulate an open-set setting we expand the representations of vehicle-fallback classes into $crane$, $excavator$, $street$-$cart$, $tractor$ and $water$-$tanker$. From the dataset, we observe that cranes are not seen on the drive-able path and so we discard this class during training. Thus our open-set setting consists of 4 novel classes.
\end{enumerate}

\subsection{Methods}
\label{approach:method}
Based on the few-shot adaptation strategy described in Section \ref{approach:prob_def} and their performance on various benchmark datasets we adopt two popular architectures.

\begin{table*}[ht]
      \caption{Roofline estimates on object detection models evaluated on IDD dataset. Faster-RCNN gives the best performance among all models.}
      \centering
      \begin{tabular}{ c | c| c | c | c }
            \hline
            Method                      & Backbone and head & $mAP$  & $mAP_{50}$ & $mAP_{75}$ \\
            \hline
            YOLO -V3\footnotemark \cite{yolo3}        & Darknet-53        & 11.7 & 26.7       & 8.9        \\
            Poly-YOLO\footnotemark[4] \cite{polyyolo}    & SE-Darknet-53     & 15.2 & 30.4       & 13.7       \\
            Mask-RCNN\footnotemark[4] \cite{maskrcnn}   & ResNet-50         & 17.5 & 30.0       & 17.7       \\
            Retina-Net \cite{retinanet}  & ResNet-50 + FPN   & 22.1 & 35.7       & 23.0       \\
            Faster-RCNN \cite{faster-rcnn} & ResNet-101 + FPN  & \textbf{27.7} & \textbf{45.4}       & \textbf{28.2}       \\
            \hline
      \end{tabular}\\
      \label{table1}
      
\end{table*}

\paragraph{Feature Similarity}
This technique, adapted from FsDet \cite{fsdet} employs non-linear similarity estimation and proposes a Two-stage Fine-tuning Architecture (TFA) which uses cosine similarity operation as an augmentation on the final classification layer (class-specific component) of Faster-RCNN.
Figure \ref{fig_tfa} illustrates the model architecture during the base and few-shot adaptation stages.
The non-linear similarity operation is introduced only in the few-shot adaptation stage as shown in Fig. \ref{fig_tfa}(b) and learns to group low-level features of a particular class and separates dissimilar features \cite{nonforget}. 
In FsDet, cosine-similarity is applied between the input features and the learnable weight vectors for each category \cite{closerfewshot} and helps reduce the intra-class variance along with prevention of catastrophic forgetting.

\paragraph{Auxiliary Network} 
This class of methods employ an additional (auxiliary) network $A$ along with a standard object detector $D$, which is trained to adapt to few-shot samples by generating class specific feature vectors.
The detector $D$ learns generalizable features from the $query$ $set$ (represented by $F_{qry}$), while the auxiliary network $A$ learns the few-shot adaptation from images in $support$ $set$ (represented by $F_{sup}$).
As shown in Fig. \ref{fig_aux}, the outputs of the detector and auxiliary network are combined to produce class specific feature sets $F_{cls}$, which is used for bounding box localization and classification. Based on the technique to produce $F_{cls}$ we evaluate two methods in our work:
\begin{itemize}
      \item First method follows Meta-Reweight \cite{reweight} and Meta-RCNN \cite{metarcnn} and performs channel-wise multiplication between $F_{sup}$ and $F_{qry}$ to generate class-specific features $F_{cls}$. This highlights the important low-level features for each class.
      \item Second method follows \cite{addfeat} (referred in the paper as Add-Info) and concatenates additional features, specifically the difference between $F_{qry}$ and $F_{sup}$, to the extended feature set $F_{cls}$. This extended feature set encodes the similarities/differences between features from various classes thus aiding in the final predictions.
\end{itemize}
Following the meta-learning paradigm, these methods also adopt episodic training strategy to adapt to few-shot data in both the base training and fine-tuning stages.
We adopt this technique as well, using a two-stage training pipeline with an additional loss term $L_{meta}$ during both training stages.
An important benefit of this technique is that it can be applied to both proposal-free and proposal-based object detection architectures as described in section \ref{related:fsod}.

\section{Experiments}
\label{experiment}
In this Section, we evaluate various few-shot techniques on IDD based on categories of data described in \ref{approach:data_split} and demonstrate the performance of the few-shot object detection networks. We also evaluate several object detection baselines that help us to establish roofline estimates for the few-shot models. 

\begin{table}[tp]
      \caption{Few-shot object detection performance ($mAP_{50}$) on IDD-10 split 1 using 10-shot samples.}
      \centering
      \begin{tabular}{c|c|c|c}
            \hline
            Method                       & $mAP$ & $mAP_{base}$ & $mAP_{novel}$ \\
            \hline
            Meta-RCNN     & 17.3  & 24.0         & 4.2   \\
            Add-Info      & 26.0  & 34.0         & 7.5   \\
            Meta-Reweight & 21.8     & 26.4       & 10.9 \\
            \textbf{TFA (FsDet)}       & \textbf{28.5}  & \textbf{31.2} & \textbf{22.1} \\
            \hline
      \end{tabular}
      \label{table3}
\end{table}
\footnotetext{Results are from \cite{polyyolo}.}

\subsection{Object Detection Baselines}
\label{experiment:od_base}
We establish a roofline (upper bound) performance by training object detection algorithms considering both proposal-free and proposal-based methods like YOLO \cite{yolo1}, Retina-Net \cite{retinanet}, Faster-RCNN \cite{faster-rcnn} on IDD dataset. The experiments generate an estimate of the performance of object detection methods given abundant data samples. The baseline evaluation is conducted on the 10 class dataset (IDD-10cls) described in section \ref{approach:data_split}.

Following state-of-the-art object detection approaches we report the mean Average Precision (mAP) in Table \ref{table1} for all methods trained on IDD (backbones are pre-trained on Imagenet). We see that Faster-RCNN based networks outperform the compared object detection networks and we consider it as the roofline for evaluating the few-shot detection task.

\subsection{Few-Shot Detection on IDD Dataset}
\label{experiment:fsod_res}
We discuss four state-of-art techniques namely, Meta-Reweight \cite{reweight}, Add-Info \cite{addfeat}, Meta-RCNN \cite{metarcnn} and FsDet \cite{fsdet} which are evaluated on representative data-splits from IDD dataset. Following previous works we report mean Average Precision at 50\% overlap of intersection over union ($mAP_{50}$) as the performance metric to evaluate the few-shot object detection architectures. All our experiments are conducted in a 10-shot setting.

\begin{table}[t]
      \caption{Few-shot object detection performance ($mAP_{50}$) on IDD-OS split using 10-shot samples.}
      \centering
      \begin{tabular}{c|c|c|c}
            \hline
            Method                   & $mAP$  & $mAP_{base}$ & $mAP_{novel}$ \\
            \hline
            Meta-RCNN & 20.1 & 27.9         & 4.2           \\
            Add-Info  & 32.0 & 33.0         & 36.0          \\
            \textbf{TFA (FsDet)} & \textbf{43.0} & \textbf{47.4} & \textbf{37.0}          \\
            \hline
      \end{tabular}
      \label{table4}
\end{table}

\begin{figure*}[t]
      \centering
      \includegraphics[height=0.19\textwidth,width=0.75\textwidth]{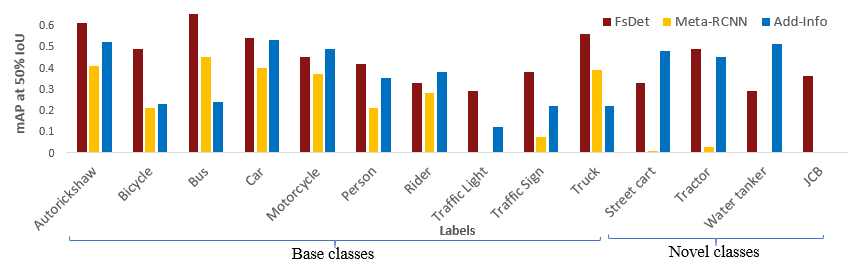}
      \caption{Classwise performance of metric-learning and meta-learning based techniques in detecting rare objects in IDD dataset. The last 4 classes represent the rare categories. FsDet, a metric learning based approach, performs better than meta-learning approaches in both base and novel classes.}
      \label{vfb_classwise}
\end{figure*}

\textbf{Implementation: } 
For meta-learning architectures (Meta-RCNN and Add-Info), we train a ResNet-101 based model for 9 epochs in the base training stage at 0.001 learning rate, followed by 9 epochs for fine-tuning. We train Meta-Reweight with a DarkNet-19 based model for 30 epochs at a 0.0001 learning rate for base training, followed by 40 epochs at 0.000001 learning-rate for finetuning.

The metric learner, FsDet, uses a ResNet-101 with Feature Pyramid Network based architecture and is trained for 40k iterations at 0.005 learning rate for the base training stage, and then followed by 15k iterations at 0.001 learning rate for fine-tuning. 

In the base training stage, all methods use a batch size of 2. All backbones are initialized with ImageNet pretrained weights and standard data augmentation like horizontal flip and random crop are applied. 

\textbf{Evaluation in Same Domain:} First, experiments are performed on IDD-10 splits containing 7 base classes and 3 novel classes per split. Table \ref{table3} reports the performance of the networks in terms of $mAP_{50}$ for all, base ($mAP_{base}$) and novel ($mAP_{novel})$ classes in the validation set of IDD. The results from split 2 (not described due to space limitations) are very similar to split 1.

\textbf{Evaluation in Open Set:} We conduct experiments on the IDD-OS split which contains 10 base classes (same used for roofline estimates) and 4 novel classes (only available in IDD). Table \ref{table4} lists the performance of all networks on IDD-OS split in terms of $mAP_{50}$ for similar class definitions mentioned above. We were not able to conduct experiments with Meta-Reweight due to memory limitations during training. Figure \ref{vfb_classwise} illustrates the classwise performance of FsDet (metric-learning) and meta-learning based architectures (Meta-RCNN and Add-Info) on IDD-OS split.

\subsection{Discussion}
\label{discussion}
Our experiments uncovered several findings on the nature of road objects and the behaviour of few-shot object detection networks in the context of driving. Results from Tables \ref{table3} and \ref{table4} demonstrate that cosine similarity based TFA architectures (FsDet) outperforms meta-learning based architectures (Meta-RCNN, Meta-Reweight and Add-Info) on novel-class performance by 11.2 $mAP$ points on IDD-10 (split 1) and 1.0 $mAP$ point on IDD-OS split. We attribute lower inter-class distance between new object categories as the probable reason for the lower performance of meta-learning over similarity-based methods. This aspect of IDD makes it a unique dataset well suited for evaluation of few-shot object detection in a real-world, driving scenarios.

 Comparing the base class performance in Table \ref{table4} against the roofline performance metrics in Table \ref{table1}, we demonstrate a lower degradation in base-class performance when adopting TFA architecture (FsDet) over its meta-learning counterparts, after the introduction of novel classes. Meta-learning techniques like Meta-RCNN and Add-Info suffer a significant reduction in base-class performance except when additional features were provided to the final prediction head of the object detector.

\begin{figure}[t]
      \centering
      \includegraphics[width=0.95\columnwidth]{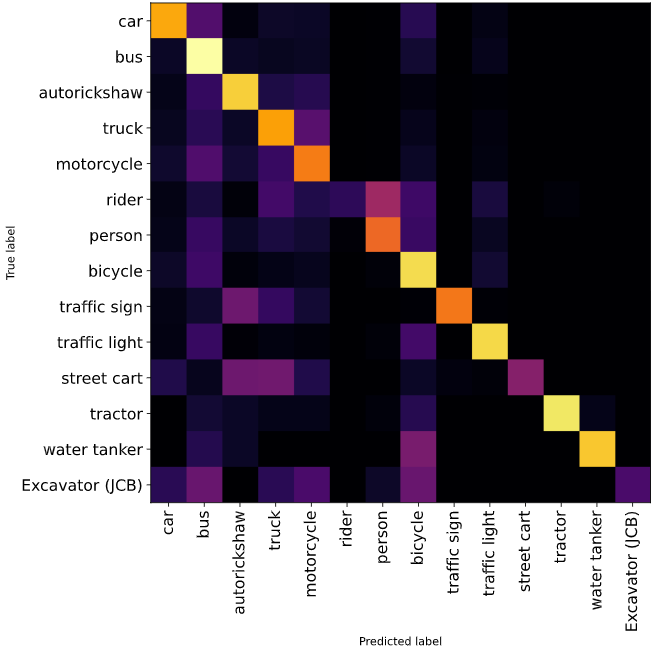}
      \caption{Confusion matrix plotted for class prediction results from IDD validation dataset showing confusion between classes when trained on IDD-OS 10-shot split on FsDet network.}
      \label{confmatrix}
\end{figure}

Figure \ref{confmatrix} shows class-level confusion among all classes in IDD-OS split trained on 10-shot data samples using TFA architecture. In particular, the confusion between $truck$ vs. $car$, $bicycle$ vs. $motorcycle$ and $water$-$tanker$ vs. $car$ classes are high, with the maximum being 40\%. This observation can possibly be explained by the fact that road objects in context, share a large number of low-level features with other object classes, thus posing a challenge for few-shot algorithms to differentiate. This observation here is in line with that by the authors of MetaDet \cite{metadet} that confusion between classes is the primary challenge in few-shot learning scenarios. This is further echoed by the authors of Meta-Reweight \cite{reweight} that there exists a high confusion of 50\% among classes in PASCAL VOC dataset.

\section{Conclusion}
We analyzed the performance of state-of-the-art methods for few-shot object detection, using a real-world dataset (IDD) which inherently contains class-imbalanced data from driving scenarios. Our evaluation of methods was for two tasks: same-domain and open-set representations. To evaluate these settings, we expanded a publicly available dataset with additional class labels in the open-set representation. By creating an extension of IDD, we hope to pave a way for many future works in few-shot learning with real-world datasets.
Based on our experiments, we conclude that cosine similarity based TFA network (FsDet) outperforms meta-learning based networks in both the tasks by 11.2 and 1.0 $mAP$ points in novel class performance respectively. We conclude that meta-learning networks while achieving great strides, tend to under perform even simpler baselines from metric-learning based methods. We also observe that class-confusions remains an open challenge in any few-shot learning paradigm and can be the focus of further improvements.

\bibliographystyle{aaai}
\bibliography{references}

\end{document}